\documentclass[journal]{IEEEtran}

\usepackage{cite}
\usepackage[numbers]{natbib}
\ifCLASSINFOpdf
\else
\fi
\usepackage{amsmath}
\usepackage{amssymb}

\usepackage{url}
\usepackage{graphicx}
\usepackage[]{subfigure}
\usepackage{siunitx}
\DeclareSIUnit\points{points}
\usepackage{tabularx}
\usepackage{adjustbox}
\usepackage{algorithm}% http://ctan.org/pkg/algorithm
\usepackage{algpseudocode}% http://ctan.org/pkg/algorithmicx

\begin{document}
% \bstctlcite{BSTcontrol}

\title{Tree Annotations in LiDAR Data Using Point Densities and Convolutional Neural Networks}
\author{Ananya~Gupta*, Jonathan~Byrne, David~Moloney, Simon~Watson, Hujun~Yin*%
\thanks{A. Gupta, S. Watson and H. Yin are with the Department of Electrical and Electronic Engineering at the University of Manchester.}%
\thanks{J. Byrne and D. Moloney are with the Intel Corporation.}
\thanks{*Corresponding Authors:\{ananya.gupta, hujun.yin\}@manchester.ac.uk}}
\maketitle

\begin{abstract}
LiDAR provides highly accurate 3D point clouds. However, data needs to be manually labelled in order to provide subsequent useful information. Manual annotation of such data is time consuming, tedious and error prone, and hence in this paper we present three automatic methods for annotating trees in LiDAR data. The first method requires high density point clouds and uses certain LiDAR data attributes for the purpose of tree identification, achieving almost 90\% accuracy. The second method uses a voxel-based 3D Convolutional Neural Network on low density LiDAR datasets and is able to identify most large trees accurately but struggles with smaller ones due to the voxelisation process. The third method is a scaled version of the PointNet++ method and works directly on outdoor point clouds and achieves an \(F_{score}\) of 82.1\% on the ISPRS benchmark dataset, comparable to the state-of-the-art methods but with increased efficiency.
\end{abstract}


% \fi
\IEEEpeerreviewmaketitle

\section{Introduction}
\IEEEPARstart{T}{rees} are essential components of both natural and urban environments; not only are they aesthetically pleasing, but also they help regulate ecological balance in the landscapes and maintain air quality by reducing particulate matter in the environment~\citep{TheNatureConservancy2016}. Urban forest inventories are important assets for planning and management of urban environments since many applications, such as mitigation of noise~\citep{Stoter2008} and creation of 3D city models~\citep{Haala2010}, make use of such data sources.  Traditional tree inventories are done manually and the process is extremely time consuming~\citep{Koch2006}.
 
Along with terrestrial LiDAR and Unmanned Aerial Vehicle photogrammetry, airborne LiDAR systems are advanced methods used for 3D data acquisition of urban environments~\citep{Schwarz2010}. LiDAR utilises laser pulses to measure distances to sufficiently opaque surfaces and objects and enable the study of 3D structure and properties of a given environment.

There is a large body of work on tree identification using LiDAR data, but most of it focuses on identifying trees in forested environments, with little emphasis given to urban environments.  Most of these methods are derivatives of the canopy height model (CHM)~\citep{Hyyppa2001,Lu2014,Reitberger2009,Smits2012,Mongus2015}.

The existing methods for tree identification in forests are not directly applicable to urban areas because the statistics of the two environments are very different. The assumption of homogeneous and highly dense collections of trees in forests does not apply in the urban environments. Urban areas are extremely complex and heterogeneous and include isolated trees and groups of trees, often of different species, ages and shapes. The presence of other vertical objects and features such as buildings and street lamps, which typically do not exist in forested environments, makes the problem even more complex.
% a standard assumption with forests is that most objects will be trees whereas with cities, there is a large presence of artificial objects such as buildings and light poles along with the natural objects.

There has been some pioneering work in urban tree detection based on machine learning. A combination of aerial images and LiDAR data has been used for segmentation followed by classification with support vector machines (SVM)~\citep{Secord2007}. This work has been extended to use features derived from  depth images of LiDAR data with a random forest classifier and achieved precision recall scores of 95\% in identifying trees in the depth images~\citep{Chen2009}. However, the accuracy degraded to below 75\% when the training and testing were done on separate datasets. Another method~\citep{Carlberg2009} used a cascade of binary classifiers to progressively identify water, ground, roofs and trees by conducting 3D shape analysis followed by region growth. Segmentation of foreground and background, followed by classification of object-like clusters using different methods such as k-nearest neighbours, SVMs and random forests was used to locate different 3D objects in an urban environment~\citep{Golovinskiy2009}. Decision trees and artificial neural networks using segmented features derived from full waveform attributes have also been used in classification~\citep{Hofle2012}.

Identification of trees in urban environments with heuristics-based methods has also been studied. \citet{Liu2013} proposed a method for extracting tree crowns by filtering out ground points and using a spoke wheel method to get tree edges. The method was able to detect over 85\% of trees from the test dataset with 95\% accuracy. However, it only focused on extracting tree crowns and did not take tree trunks into account and was unsuited for urban forest inventory applications. A voxel-based method was used to extract individual trees from mobile laser scanning data but it was not suitable for use with airborne LiDAR scans ~\citep{Wu2013}. A combination of LiDAR and hyperspectral data to detect treetops and a region growing algorithm for segmentation has also been developed for urban forest inventory purposes~\citep{Zhang2015}. 

Recent work based on deep learning has shown promising results in identifying objects in LiDAR scans of urban environments. \citet{Yousefhussien2018} used a 1D convolutional neural network (CNN) in conjunction with LiDAR data and spectral information to generate point-wise semantic labels for unordered points and achieved a mean \(F_{score}\) of 63.32\% on the ISPRS Benchmark~\citep{Rottensteiner2014}. Multiple CNNs have been used to learn per-point features of different data attributes (height, intensity and roughness) from multi-scale images for classification~\citep{Zhao2018}. However, this method was computationally expensive as it generated multiple contextual images per point in the dataset and classified each point individually. Another method along similar lines used a CNN in conjunction with 2D images derived from the point clouds for point-based classification~\citep{Yang2017a}. The method was also extremely computationally expensive and relied on hand-engineered feature images, contrary to the popular use of CNNs for extracting features implicitly. The method also required a cleanly labelled training set with all categories for multi-category classification.
% \citet{Niemeyer2016} \todo{add description}

In this paper, we propose three methods for automatic tree identification in LiDAR data in urban environments. All three methods work on point cloud data and the third method has been adapted to combine spectral data with the point locations.

% and scales the state-of-the-art method in indoor point cloud segmentation for outdoor datasets.
% as compared to the combination of point clouds and hyperspectral images required by the current state-of-the-art methods.~\todo{correct this}

\begin{figure}[]
\centering
\subfigure[Low Density Point Cloud.]{\includegraphics[width=.23\textwidth]{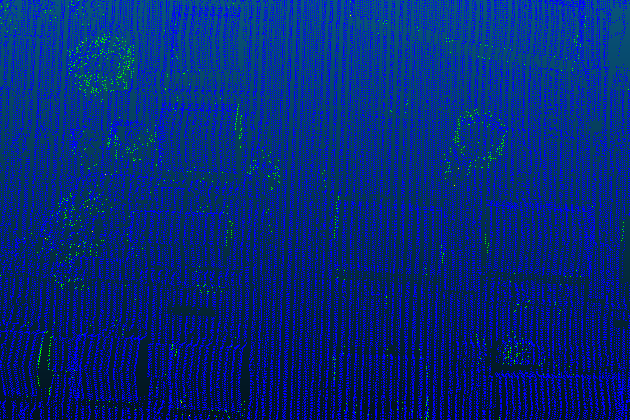}}
\subfigure[High Density Point Cloud.]{\includegraphics[width=.23\textwidth]{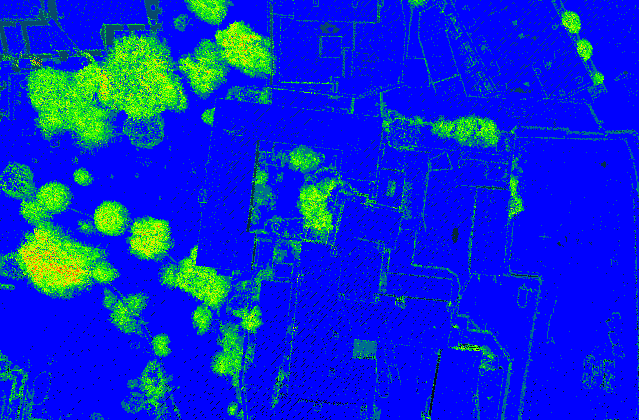}}
\caption{Point clouds with red showing high number of returns decreasing to blue representing single return. \textbf{Left}: Low density point cloud hence very few points with high number of returns per point. \textbf{Right}: High density point cloud hence more dense collections of points with a high number or returns per point.}
\label{fig:pc_density}
\end{figure}

The first method, termed as MultiReturn, is based on our earlier work~\citep{Gupta2018} and uses traditional handcrafted features as well as inherent data characteristics of LiDAR data. It works well on datasets that have point cloud density \(>\)\SI{20}{\points\per\metre\squared} as these datasets contain the \textit{number of returns} characteristic that allows the method to identify trees. However, it is unable to deal with point clouds with lower resolutions, since they do not exhibit the same characteristic, as illustrated in Figure \ref{fig:pc_density}.

The second method, termed as TreeNet, is based on a 3D CNN and works on voxelised datasets. It can be trained using the results of the first method and is able to identify large trees with good accuracy. However, it struggles to identify small trees due to the limited resolution of voxels.

The third method, sPointNet++, is based on PointNet++~\citep{Qi2017a}, a state-of-the-art method for 3D shape classification and indoor point cloud segmentation. We scale it to include spectral information for dealing with outdoor aerial datasets, which are much noisier than indoor data. It works directly with point clouds and is able to identify smaller trees  well. However, it requires a large amount of cleanly labelled training data, which can be troublesome to obtain. We adapt the training loss to deal with unbalanced class distributions,  allowing the network to train a binary classifier with very few positive points (tree category) relative to large negative points (non-tree points).

\section{Methods}

%\IEEEpubidadjcol

\subsection{MultiReturn: Tree Annotation with Number of Returns}

The first method, \textit{MultiReturn}~\citep{Gupta2018}, which can be reformulated in Algorithm \ref{alg:tree_ret}, is based on four distinct steps:
\begin{itemize}
\item Ground filtering
\item Voxelising non-ground point cloud data
\item Isolating tree-like regions using the information gained from the number of returns
\item Post-processing to remove false positives.
\end{itemize}

\begin{algorithm}
\caption{Tree Annotation with Number of Returns}\label{alg:tree_ret}
\hspace*{\algorithmicindent} \textbf{Input}: Point Cloud \textit{pcd} 
\hspace*{\algorithmicindent} \textbf{Output}: Tree data \textit{trees} 

\begin{algorithmic}[1]
    \State $filtered\_cloud$ = []
    \State $filtered\_vox$ = []
    \State $trees$ = []
    \State $gnd\_points$ = PMF($pcd$)
    \For{$point$ in $pcd$}
        \If{$point$ not in $gnd\_points$}
            \State $filtered\_cloud$.append($point$)
        \EndIf
    \EndFor
    \State $vox\_cloud$ = Voxelise($filtered\_cloud$)
    \For{$vox$ in $vox\_cloud$}
        \If{$vox$.$no\_of\_returns$ $>$ $ret\_thresh$}
            \State  $filtered\_vox$.append($vox$)
        \EndIf
    \EndFor
    \State $vox\_regions$ = ConnectedComponents($filtered\_vox$)
    \For{$region$ in $vox\_regions$}
        \If{$region$.size $>$ $comp\_threshold$}
            \If{$region.x$/$region.y$ $\geq$ 2 or $region.y$/$region.x$ $\geq$ 2} 
                \State continue;
            \Else
                \State$trees$.append($region$)
            \EndIf
        \EndIf
    \EndFor

\end{algorithmic}
\end{algorithm}

\begin{figure}[]
\centering
\subfigure[Original Point Cloud.]{\includegraphics[width=.23\textwidth]{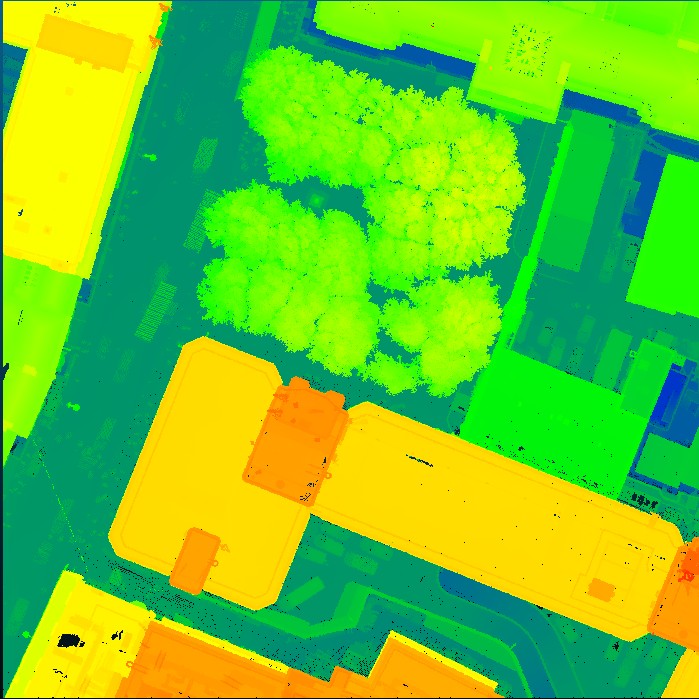}}\hfill
% \subfigure[Result of filtering with PMF]{\includegraphics[width=.3\textwidth]{images/gnd1.jpg}}
\subfigure[Result after cleaning for ground and  noise.]{\includegraphics[width=.23\textwidth]{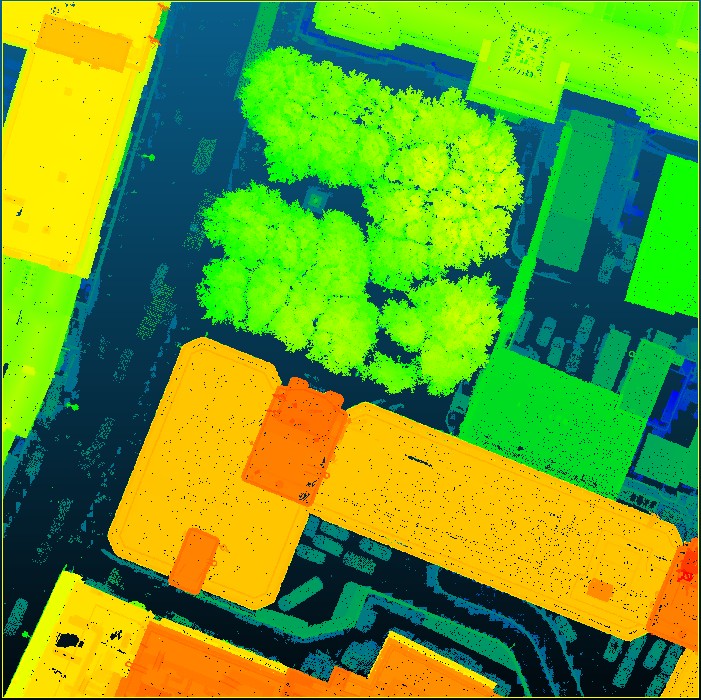}}
\caption{Ground filtering process with points coloured according to height.}
\label{fig:ground_filter}
\end{figure}

In order to identify and filter the ground points, we use a progressive morphological filter (PMF)~\citep{Zhang2003}. It uses  morphological erosion and dilation operations in conjunction with windows of progressively increasing size to identify non-ground points. This is followed by statistical outlier removal to remove noisy points that are below the ground level and a filtered result can be seen in Figure \ref{fig:ground_filter}(b). 

The publicly available Point Data Abstraction Library (PDAL)~\citep{pdal} was used for this filtering process. The \textit{filters.pmf} package with a window size of \SI{40}{\metre} and maximum distance of \SI{3.5}{\metre} was used for ground filtering and the \textit{filters.outlier} function was used for statistical outlier removal with the default values given in the package.

% As can be seen from the results of the filtering process in Figure \ref{fig:ground_filter}(b), there are still some ground points left in the output. Hence we do statistical outlier removal and much cleaner results can be obtained as seen in Figure \ref{fig:ground_filter}(c).

The point cloud is converted into a volumetric occupancy grid in order to make the data easier to deal with, because it reduces dimensionality. A fixed size 3D grid is overlaid on the point cloud and the occupancy of each cell depends on the presence of points within the cell, i.e. the cell is unoccupied if there are no points in the cell's volume and vice versa. In this case, each volumetric element, voxel, represents a grid cell in the subsampled point cloud.

% We convert the data tiles of \(100m \times 100m\) into a voxel grid of size  \(256\times256\times256\) with a voxel resolution of \(\approx 0.39m\times0.39m\times0.39m\). Any further increase in resolution does not noticeably increase accuracy and causes an exponential increase in the processing time due to the increased dimensionality of the data.

We use VOLA~\citep{Byrne2018} to encode the voxel representation in a sparse format. VOLA is a hierarchical 3D data structure that draws inspiration from octree based data structures. But unlike standard octrees it only encodes occupied voxels in a ``1-bit per voxel" format, and hence is extremely memory efficient. In this case we use a ``2-bit per voxel" format to encode additional information per voxel such as colour, number of returns and intensity value.

LiDAR datasets are acquired by pulsing laser light and measuring the time that the pulse takes to reflect from sufficiently opaque surfaces to calculate the distance to said surface. A single pulse can reflect completely in one collision with a surface or can reflect multiple times when it encounters edges that reflect the light partially. In high point density datasets, trees typically have multiple returns as pulses are partially reflected from the edges of leaves. Leaf-off trees also have similar characteristics since they still have a number of branches and twigs giving similar patterns of returns. Other features that can have a high number of returns are building edges and window ledges. However, returns in these latter cases are more scattered than that in the case of trees, where a large number of high returns are closely packed as can be seen in Figure \ref{fig:pc_density}(b).

\begin{figure}[]
\centering
\includegraphics[width=0.4\textwidth]{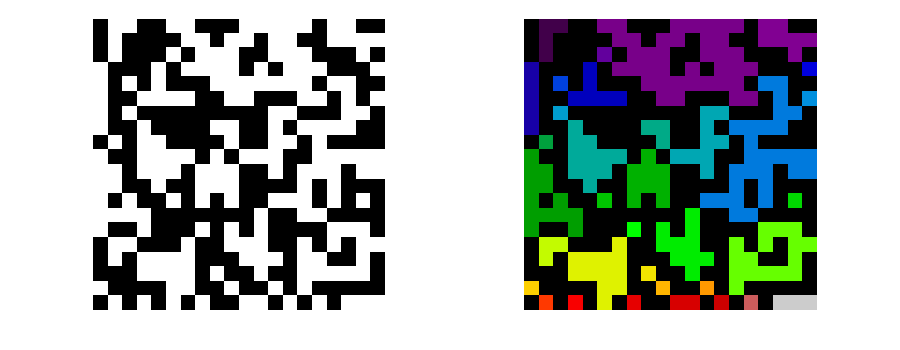}
\caption{Connected Components in 2D. \textbf{Left}: White indicates occupied cells, Black unoccupied; \textbf{Right}: Each region identified as a connected block is represented by a different colour.}
\label{fig:conn_components}
\end{figure}

We use this insight to isolate tree regions by identifying voxels with multiple returns (empirically identified as more than 3 returns per voxel) and then performing a connected component analysis on these voxels. A connected component is a subgraph in an undirected graph where any two vertices are connected to each other by paths. In this case, regions of dense voxels with at least one common edge are marked as connected components. An example of 2D connected component labelling has been shown in Figure \ref{fig:conn_components}, this is easily extendable to 3D by replacing the 2D cells with 3D voxels.

Regions of high returns with a minimum number of connected components, \(ncc\), are then identified as tree regions, while any regions with fewer components than the threshold value are discarded as noise from buildings, corners, etc.

\begin{equation}
f_x = \left\{\begin{matrix}
1, & ncc \geq threshold\\ 
0 & ncc < threshold
\end{matrix}\right\}
\end{equation}

The tree regions isolated by connected components are typically tree canopies as trunk voxels with a high number of returns are typically sparse and not well connected. In order to find tree trunks, the maximum and minimum \(x\) and \(y\) coordinates of each region are identified, along with the maximum \(z\) coordinate. These coordinates are then used to place a 3D bounding box in the original data near the ground level in order to capture the trunk information. The width to length ratio of the bounding box is constrained so that they are approximately equal. Any regions not matching these constraints are discarded as a false positive. Upon manual inspection of the results and statistical analysis, it could be seen that although this resulted in removing some trees, all the walls covered with ivy are removed.

% Upon manual inspection of the results, it could be seen that this removed some trees, especially those with merged canopies but it was required to remove walls covered with ivy as these walls are typically long but not very thick.

% and hence the width to length ratio \(\approx1\). 

% This method is applicable for high density point cloud datasets \(>\)\SI{30}{\points\per\metre\squared} because these datasets contain the high number of returns characteristics that allows our method to identify trees. However, it is unsuitable for point clouds with lower resolution since they do not exhibit the same characteristics as shown in Figure \ref{fig:pc_density}.

\subsection{TreeNet: 3D CNN for Tree Segmentation}

\begin{figure*}[]
\centering
\includegraphics[width=0.95\textwidth]{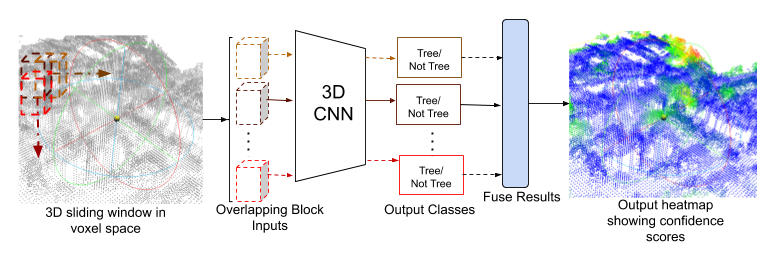}
\caption{Block Diagram for TreeNet Segmentation. \textbf{Left}: LiDAR point cloud subdivided into overlapping blocks. \textbf{Middle}: Each block processed individually by CNN with binary output. \textbf{Right}: Results fused to provide a confidence score per point.}
\label{fig:3d_cnn}
\end{figure*}

Inspired by the recent successes of CNNs in image classification, we propose \textit{TreeNet}, a deep learning approach based on 3D voxels to deal with low density point clouds. A 3D CNN is designed for binary classification of 3D voxel spaces for presence or absence of trees. The output labels from sliding voxel windows of the network are fused to provide a per voxel confidence score. The system architecture is shown in Figure \ref{fig:3d_cnn}.

\begin{table}[h]
\caption{3D CNN Architecture Details}
\label{tab:3d_cnn} \centering
\begin{tabular}{|c|c|}
  \hline
  \textbf{Layer Type} & \textbf{Details}\\
  \hline
  Conv3D & 5x5x5 filter size\\
  & 32 filters \\
  & stride = 2 \\
  \hline
  Dropout3D & \textit{p} = 0.2 \\
  \hline
  Conv3D & 3x3x3 filter size \\
  & 32 filters\\
  & stride = 1 \\
  \hline
  MaxPool3D & 2x2x2 window size  \\
  \hline
  Dropout3D & \textit{p} = 0.3 \\
  \hline
  FC & 128 features\\
  \hline 
  Dropout3D & \textit{p} = 0.4 \\
  \hline
  FC &  2\\

  \hline
\end{tabular}
\end{table}

3D CNNs are extensions of the standard 2D CNNs and are used here for two main reasons: their ability to learn local spatial features in 3D space instead of relying on traditional handcrafted features, and their ability to encode more complex relationships of hierarchical features with combinations of multiple layers. The input to the network is a multi-channel 3D volume in \(\mathbb{R}^{ w \times h \times d \times c_i}\) where \(w\), \(h\) and \(d\) are the spatial dimensions of the input volume and \(c_i\) is the number of input channels.  

The data is processed through a series of 3D convolutional layers. Each layer, \(l\), consists of \(c_l\) filters where each 3D filter, \(K^l\), is convolved with the input voxel (for the first layer) or the output of the previous layer, \(O^{l-1}\) (for subsequent layers). The output from layer \(l\) is given by \(O^l\) and is calculated using Equation \ref{eq:conv}.

% . Each layer \(l\) consists of \(c_l\) filters that are convolved with the input (for the first layer) and the output of the previous layer, \(O^{l-1}\) (for subsequent layers). 

% of size \(f_1 \times f_2 \times f_3 \times c_{l-1}\), where \(f_{1,2,3}\) gives the spatial dimension of the filter. These kernels are convolved with the input (for the first layer) and the output of the previous layer (for subsequent layers) to output \(c_l\) feature maps. The mathematical formulation for this process is given in Equation \ref{eq:conv} where. 

% \begin{equation}
% \begin{multline}
% \label{eq:conv}
%  O(i, j, k, c_l) = \\ \sum_{c_{l-1},f_1, f_2,f_3} I(c_{l-1},i+f_1, j+f_2, k+f_3) K(c_{l-1},f_1,f_2, f_3, c_l)  
% \end{multline}
% \end{equation}

\begin{equation}
\label{eq:conv}
O^l_n = \sum_{m=0}^{c_{l-1}} O_m^{l-1}* K^l_{m,n} \textrm{ where } n=1,2,3, ...,c_l
\end{equation}

The output activations from the convolutional layers are passed through a leaky Rectified Linear Unit (ReLU)~\cite{Nair2010, Maas2013} with a slope of 0.1. A 3D Max Pooling layer is used for downsampling the data, hence reducing the dimensionality and making the computation more efficient. Dropout~\cite{Srivastava2014} layers are used for preventing overfitting. 

The final two layers in the network are fully connected (FC) layers to learn weighted combinations of the extracted feature maps. The cross entropy loss is used for training the  network. 

The details of the CNN architecture are listed in Table \ref{tab:3d_cnn} where the input is a 20x20x100 voxel volume. Variable \textit{p} in the dropout layer indicates the probability of the elements being replaced with zeroes. 

The output from the final FC layer is a binary class label indicating whether the input box area contains a tree or not. A sliding window with overlap in both horizontal dimensions is used over the dataset for classification. In order to convert this to a per-voxel result, the output classification is mapped back to the input voxels. With the overlapping windows used, each voxel has multiple output values. A voting scheme is used to get the final result. If over 40\% of these outputs are positive, the voxel is identified as belonging to a tree.  The 40\% threshold was chosen empirically based on the experimental results.

\subsection{Scaling Pointnet++}
%Since the voxel based 3D CNN is unable to identify small trees due to the decreased resolution in the voxelisation process,

PointNet++~\citep{Qi2017a} is the state-of-the-art method in point cloud segmentation of indoor environments. Herein, we provide a scaled version called \textit{sPointNet++} to deal with aerial urban LiDAR scans and to work directly with unordered point sets instead of the regularly spaced voxel grids as in the previous methods. Aerial laser scans, in comparison to indoor scenes, tend to be noisier and have more terrain and point density variations across scenes. In order to deal with these issues, we augment the model with the use of spectral information in addition to point cloud data.

The 2D spectral aerial image containing IR-R-G (Infrared-Red-Green) values is fused with the point cloud. Bilinear interpolation in the image plane is used to assign IR-R-G information to each point in the dataset. The outline of the model used is given in Table \ref{tab:pointnet_arch} and the description of the layers is as follows:

\textit{Sampling and Grouping Layer}: uses furthest point sampling (FPS)~\cite{Moenning2003}, which is an iterative sampling process that picks the next sample from the least known region in the sampling domain, to identify a subset of \(N_l\) input points as centroids. These centroids are used to identify points within a local region around the centroids using a ball query of radius \(r\) and the output from this layer are point sets of size \(N_l \times K \times \left(d + C\right)\), where \(K\) is the number of points in a local region around each centroid, \(d\) is the dimensionality of the point coordinates, and \(C\) is the dimensionality of point information such as spectral data. This allows for uniform coverage of point clouds that have non-uniform point density, common in aerial laser scans due to varying scanning patterns. In order to learn hierarchical features, a number of these layers are stacked in a series with varying radius for the neighbourhood ball query to encode features at different resolutions.

\textit{PointNet layer}: the output from the sampling and grouping layer is processed with a PointNet layer~\citep{Qi2017} which learns an abstracted local feature per centroid. The output data has size  \(N_l \times C'\), where \(C'\) gives the abstracted features per centroid. The PointNet layer is a set of \(n\) 1x1 convolution layers with the numbers of filters in each 1x1 layer given by \([l_1,..,l_n]\).

\begin{figure}[]
\centering
\includegraphics[width=0.5\textwidth]{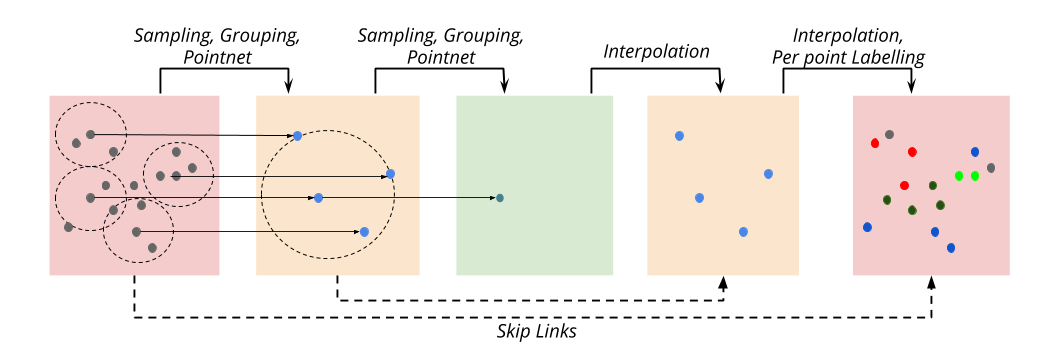}
\caption{sPointNet++ Architecture.}
\label{fig:pointnet_pp_arch}
\end{figure}

\textit{Feature Set Propagation}: in order to provide a per point label for semantic segmentation, the feature labels are propagated from the subsampled points to the original points with distance-based interpolation and skip links as shown in Figure \ref{fig:pointnet_pp_arch}. Each point could be used as a centroid to get per point labels but only at the cost of high computation. The interpolation is done with a less computationally intensive method explained below.

The feature propagation layers are essentially a mirror of the feature sampling layers. In set sampling, the number of input points and output points is \(N_{l-1}\) and \(N_l\) respectively, where \(N_l < N_{l-1} \), whereas in feature propagation the input is \(N_l\) and the output is \(N_{l-1}\). This is done by interpolating the \(N_l\) values to \(N_{l-1}\) using an inverse-distance weighted average of \(k\) nearest neighbours, with \(k\) being set to 3 and the weight being given by the inverse of the Euclidean distance between the points. These are concatenated with skip link features from the corresponding sampling layer followed by \(m\) 1x1 convolution layers, with the width of each 1x1 layer given by \([l_1,..l_m]\) in Table \ref{tab:pointnet_arch}. The final layers are a number of FC layers (implemented as 1x1 convolutions to allow for variable size inputs) with ReLU activations. The output is a per-class confidence score for every input point. 

% The network gives a class label per input point.

\begin{table}[]
\caption{sPointNet++ Architecture}
\begin{tabular}{|l|l|}
\hline
\textbf{Layer} & \textbf{Details}  \\ \hline
Sampling and Grouping & \(N_1\)=1024, r=1.5 , K=32 ,d=3, C=3) \\ \hline
PointNet & [32,32,64] \\ \hline
Sampling and Grouping &   \(N_1\)=256, r=3 , K=32) \\ \hline
PointNet &  [64,64,128] \\ \hline
Sampling and Grouping &   \(N_1\)=64, r= 6 , K=32) \\ \hline
PointNet &  [128,128,256] \\ \hline
Sampling and Grouping &   \(N_1\)=16, r=12 , K=32) \\ \hline
PointNet &  [256, 256, 512] \\ \hline
Feature Set Propagation & [256,256]  \\ \hline
Feature Set Propagation & [256,256] \\ \hline
Feature Set Propagation & [256,128] \\ \hline
Feature Set Propagation & [128,128,128] \\ \hline
FC & [128] \\ \hline
FC & [2] \\ \hline
\end{tabular}
\label{tab:pointnet_arch}
\end{table}

Since the class distribution is highly unbalanced with only about 15\% of the points in the training set being positive samples, we use a weighted cross entropy loss to train the network. The output classes are weighted inversely to the occurrence of a specific label in the training set, so that classes which occur less frequently are weighted more when calculating the output loss.  This discourages the network from always labelling the few positive samples as negative since this is a local minima in the unweighted loss space that the network would converge to. The loss function used is:
\begin{equation}
L = -  w_c \left[log \frac{e^{o_c}}{\sum_{j=1}^{C}e^{o_j}}\right]
\end{equation}
which can be rewritten as:
\begin{equation}
L =w_c \left[-o_c + log \sum_{j=1}^{C}e^{o_j} \right ]
\end{equation}
where \(L\) is the loss per sample, \(o_c\) is the output value for the target class \(c\) and \(w_c\) is the weight of the target class which is calculated over all classes \(C\) as:
\begin{equation}
w_c = \frac{\max \limits_{i \in \{1,..,C\}} \left( S_i\right)}{S_c}
\end{equation}
where \(S_i\) is the number of samples of class \(i\).

In order to train the network, random blocks of different sizes were sampled from the point cloud as training samples, with 4096 sampling points per block. The sampling ball radius was varied between \SI{0.3}{\metre} and \SI{2}{\metre}, while the block width varied between \SI{10}{\metre} to \SI{20}{\metre} for different data types. The block size informs the encoding  of the global features while the sampling ball radius affects the local features learnt by the network. However, due to the inherent multiscale feature encoding in the network, it was seen that the results remained mostly consistent across different sizes and hence the results reported were obtained with the input sampling radius of \SI{1.5}{\metre} and block size of \SI{15}{\metre}.

\section{Experimental Setup}
\subsection{Datasets}
\label{sec:datasets}
The following datasets were used for evaluating the proposed methods:
\begin{itemize}
\item 2015 Aerial Laser and Photogrammetry Survey of Dublin City~\citep{Laefer2015s}
\item Montreal 2015 Aerien Survey~\citep{montreal2015}
\item ISPRS Urban Classification Benchmark dataset~\citep{Rottensteiner2014}
\end{itemize}

The MultiReturn method was tested on the Dublin city dataset. This dataset, captured at an altitude of \SI{300}{\metre} using a TopEye system S/N 443, consists of over 600 million points with an average point density of \SI{348.43}{\points\per\metre\squared}. It covered an area of \SI{2}{\kilo\metre\squared} in Dublin city centre. Following the original tile dimensions of \(\SI{100}{\metre}\times\SI{100}{\metre}\) of the dataset, each tile was converted into a voxel grid of dimensionality \(256\times256\times256\) hence limiting the resolution of the voxel grid to \(\approx \SI{0.39}{\metre}\times\SI{0.39}{\metre}\times\SI{0.39}{\metre}\) per voxel. 
% Any further increase in resolution would not noticeably increase accuracy but cause an exponential increase in the processing time due to increased dimensionality of the data.

The results from the MultiReturn algorithm were validated using the labels from \citet{Ningal2012} containing tree annotations around some of the major streets in Dublin from 2008. In order to get more up to date results, the region north of the Liffey river was manually annotated for trees using satellite imagery from Google Earth in 2015.

The Montreal dataset is an aerial survey of the territory of Montreal city. It covers an area of over \SI{775}{\kilo\metre\squared}, large enough for use in CNN training. The MultiReturn method was applied on this dataset for labelling trees which were used as positive samples for training TreeNet. An equal number of negative samples were sampled randomly from this dataset.

% Since the Dublin dataset is a small area with a very limited number of trees, insufficient amount of data for training the CNN, the Montreal dataset was used for identifying individual trees. An equal number of negative samples were sampled randomly from the training dataset. This data was then used to train the CNN for the second task of segmenting tree voxels. 

The ISPRS dataset was acquired using the Leica ALS50 system and has a point density of approximately \SI{8}{\points\per\metre\squared}. It has been provided by the German Society for Photogrammetry, Remote Sensing and Geoinformation (DGPF) over Vaihingen in Germany~\citep{Cramer2010}. It has per point class labels and separate training and test sets, which were used to train sPointNet++ and test the segmentation results.

In all cases for training the CNNs, 10\% of the training dataset was held aside as a validation set. 

\textit{Note on Point Density:} most LiDAR datasets report average point density as a statistic of the dataset. However, this statistic is not always well defined and can vary across datasets due to intentional or unintentional bias~\citep{Naus2006}. Furthermore,  a small LiDAR dataset can have a large variance in its point density and due to the law of large  numbers, the variance decreases as the dataset becomes larger (millions or billions of points).

% A number of factors can affect the point spacing and density of the dataset
% \begin{itemize}
% \item Scanning pattern
% \item Scanning hardware
% \item Differences in ground structure: a flat surface is likely to have a more uniform distribution as compared to areas with buildings, trees and varying terrain.
% \end{itemize}

\begin{figure}[]
\centering
\subfigure[Vaihingen Point Density.]{\includegraphics[width=0.3\textwidth]{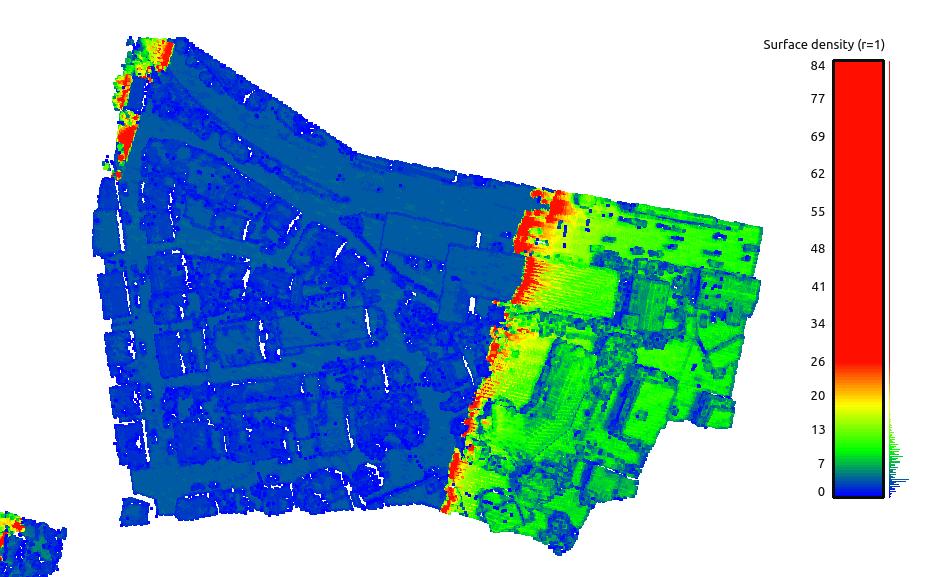}}\hfill
\subfigure[Montreal Point Density.]{\includegraphics[width=0.3\textwidth]{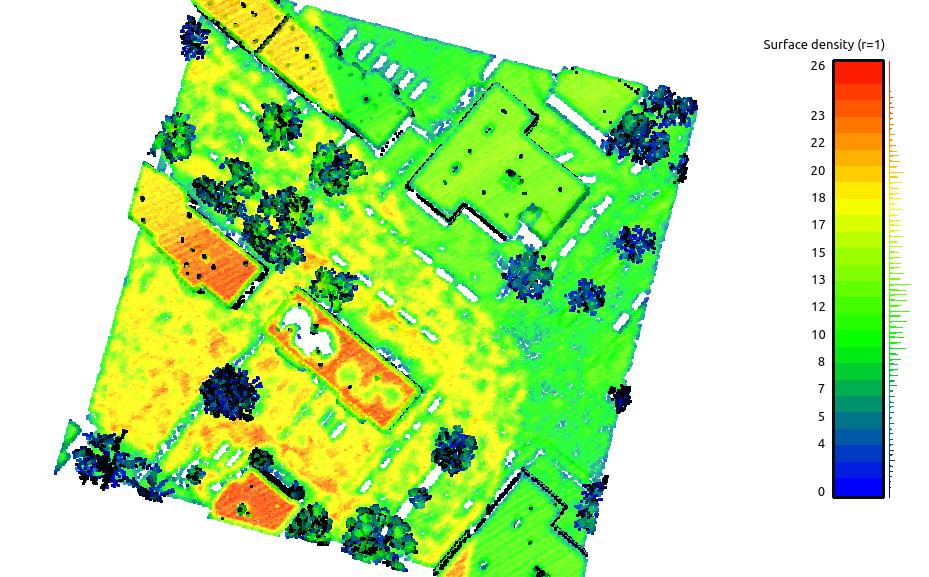}}
\caption{Surface Density plotted as heatmap for different LiDAR datasets. The density range was different between the two datasets and hence the heatmap colour scale was saturated at the same value to enable a valid visual comparison.}
\label{fig:point_density}
\end{figure}

Point density also varies within a dataset and there is no standard metric to calculate point density across datasets. Some may report theoretical density values whereas others calculate them from on the dataset. Hence, even though both the Montreal and the ISPRS datasets reportedly have a point density of \SI{8}{\points\per\metre\squared}, their actual point statistics are quite different as can be seen in Figure \ref{fig:point_density} where the surface point densities of the two datasets are plotted as a heatmap. The figure shows that the point density varies across the two datasets and within the same datasets, especially in the case of Vaihingen, which is extremely sparse in parts of the scan. Due to this disparity, only the Montreal dataset could be used in the first method for tree annotation.

\subsection{Evaluation Metrics}

The metrics presented in the ISPRS 3D labelling contest were used in the experiments for the purpose of evaluation. These metrics are precision or correctness, recall or completeness and \(F_{score}\), defined as follows:

\begin{equation}
\begin{split}
r=\frac{TP}{TP+FN}  \\
p=\frac{TP}{TP+FP} \\
F_{score} = 2\times \frac{r\times p}{r + p}
\end{split}
\end{equation}
where \(TP\) is the number of true positives, \(FN\) is the number of false negatives, \(FP\) is the number of false positives and \(F_{score}\) is the overall accuracy.
 
For the purpose of MultiReturn, the results were based on the detection of individual trees where a tree was assumed to be correctly identified if the predicted stem location was within \SI{1.5}{\metre} of the actual stem location. The results for TreeNet and sPointNet++ were evaluated per voxel and per point, respectively.

\subsection{Tools}

The initial noise and ground filtering of the LiDAR dataset was done using the publicly available Point Data Abstraction Library (PDAL)~\citep{pdal}. CloudCompare~\citep{Cloudcompare} was used for the point cloud visualisations. PyTorch~\citep{pytorch} and TensorFlow~\citep{Abadi2015} were used for building and training the CNNs.

\subsection{Training Details}

TreeNet was trained using stochastic gradient descent with an initial learning rate of 0.01, momentum of 0.9 and exponential weight decay at a rate of 0.001. It was trained for up to 50 epochs.

sPointNet++ was trained using the Adam optimiser~\citep{Kingma2014} with an initial learning rate of 0.001 and exponential decay at a rate of 0.7. It was trained till the validation loss converged up to a maximum of 200 epochs.

\section{Results And Discussion}

\subsection{Tree Annotation with MultiReturn}

\begin{figure}[]
\centering
\subfigure[Experiment 1: Results compared to labels from manual study done in 2008.]{\includegraphics[width=0.4\textwidth]{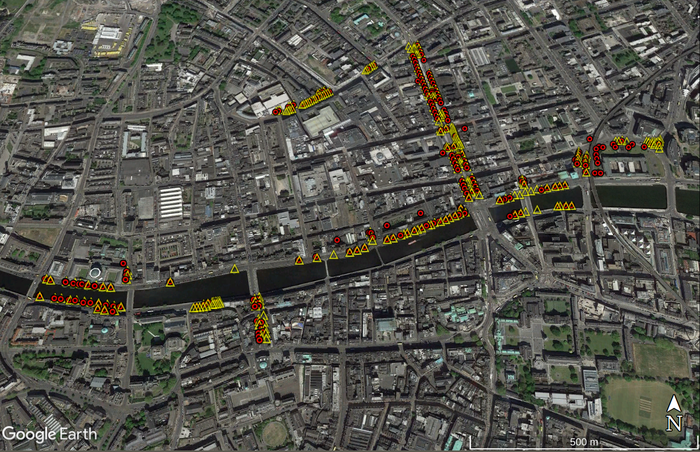}}\hfill
\subfigure[Experiment 2: Results compared to labels from Google Earth Imagery in 2015. ]{\includegraphics[width=0.4\textwidth]{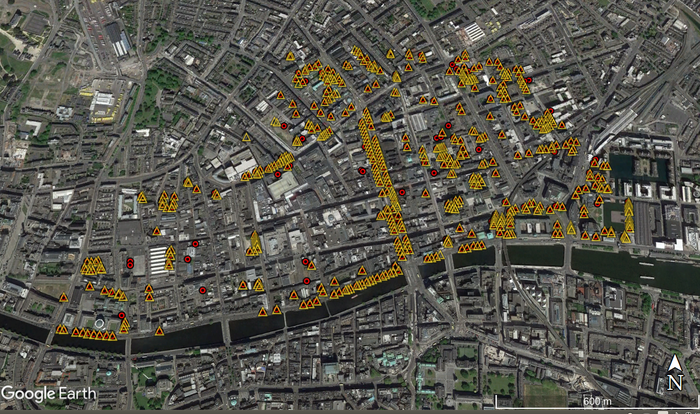}}
\caption{Map of survey area with tree locations shown in yellow and the outputs of our MultiReturn~\cite{Gupta2018} labelling algorithm shown in red.}
\label{fig:det_trees}
\end{figure}

Locations of trees in the Dublin dataset labelled by the MultiReturn method are shown in Figure \ref{fig:det_trees} along with two sets of manually labelled locations as explained in Section \ref{sec:datasets}. The performances are summarised in Table \ref{tab:anno_res}. 

The results from Experiment 1 seem to suggest that the accuracy of this labelling method is not very high with an \(F_{score}\) of 0.66. However, since the original annotations for this experiment were from 2008, it was discovered that they were out-of-date since the urban landscape of the city had changed significantly between the dates of the annotations and the acquisition of the LiDAR dataset for the development of the city tram network. 

\begin{table}[h]
\caption{Summary of Results using MultiReturn}\label{tab:anno_res} \centering
\begin{tabular}{|c|c|c|c|c|c|c|c|}
  \hline
  \textbf{Experiment} & \textbf{Trees} & \textbf{TP} & \textbf{FP} & \textbf{FN} & \textbf{\(p\)} & \textbf{\(r\)} & \textbf{F\(_{score}\)}  \\
  \hline
  1 &  313 & 178 & 45 & 135 & 0.57 & 0.8 & 0.66\\
  \hline
  2 & 535 & 469 & 56 & 66 & 0.88 & 0.89 & 0.88 \\
  \hline
\end{tabular}
\end{table}

Experiment 2 used annotations from the aerial imagery of 2015 acquired from the Google Earth for a fair comparison. The proposed method gave good results with an \(F_{score}\) of 0.89. It was especially good in identifying isolated trees with non-converging canopies. However, it did not perform well in areas such as parks where the tree canopies were merged as it could not identify such trees individually. Some merged canopies were removed by the tree width and height constraints, which were required for the removal of ivy-covered walls.

\subsection{TreeNet for Segmentation}

The ISPRS test dataset was voxelised to a resolution of \(\approx \SI{0.39}{\metre}\times\SI{0.39}{\metre}\times\SI{0.39}{\metre}\) per voxel to match the resolution of the Dublin dataset. From the statistics of the training dataset, it was observed that most of the trees would have spatial dimensions of less than \SI{8}{\metre} \(\times\) \SI{8}{\metre} with height varying upto \SI{25}{\metre} and hence the input voxel space for classification as set as 20 vox \(\times\) 20 vox \(\times\) 100 vox. 

% The New York city LiDAR dataset was also used for training a tree classifier. However, since this dataset isn't as dense as the Dublin dataset and has a nominal pulse spacing of 0.7m, it can only be voxelised to a depth of 3 with voxel resolution \(\approx 1.6m\times 1.6m\times1.6m\), if it is voxelised any further, the voxels become very disconnected. In this case, the test dataset was also voxelised to the same resolution.

\begin{table*}[]
\centering
\caption{TreeNet Segmentation results on ISPRS test dataset for different parameters}
\label{tab:aerial_seg}
\begin{tabular}{|c|c|c|c|c|c|c|c|}
\hline
\begin{tabular}[c]{@{}c@{}}\textbf{Training} \\\textbf{Dataset} \end{tabular} &
\begin{tabular}[c]{@{}c@{}}\textbf{Val} \\\textbf{Accuracy \%}\end{tabular} &
\begin{tabular}[c]{@{}c@{}}\textbf{Corrupted} \\\textbf{Data}\end{tabular} &
\begin{tabular}[c]{@{}c@{}}\textbf{Min Neg} \\\textbf{Components}\end{tabular} &  \textbf{\(p\)} & \textbf{\(r\)} & \textbf{F\(_{score}\)} \\ \hline
Dublin &  93.26 & False & 5 & 0.51 & 0.11 &0.18 \\ \hline
Montreal & 91.81 & False & 20 &0.54 & 0.6 & 0.57 \\ \hline
Montreal & 92.69 & True & 20 & 0.57 & 0.62 & 0.59 \\ \hline
Montreal & 91.20 & True & 5  & \textbf{0.6} & \textbf{0.64} &\textbf{ 0.62} \\ \hline

ISPRS Training & 74.43 & True & 5 & 0.21 & 0.15 & 0.17 \\ \hline
\end{tabular}

\end{table*}

\begin{figure}[]
\centering
\subfigure[Ground Truth- Tree voxels shown in Red.]{\includegraphics[width=0.23\textwidth]{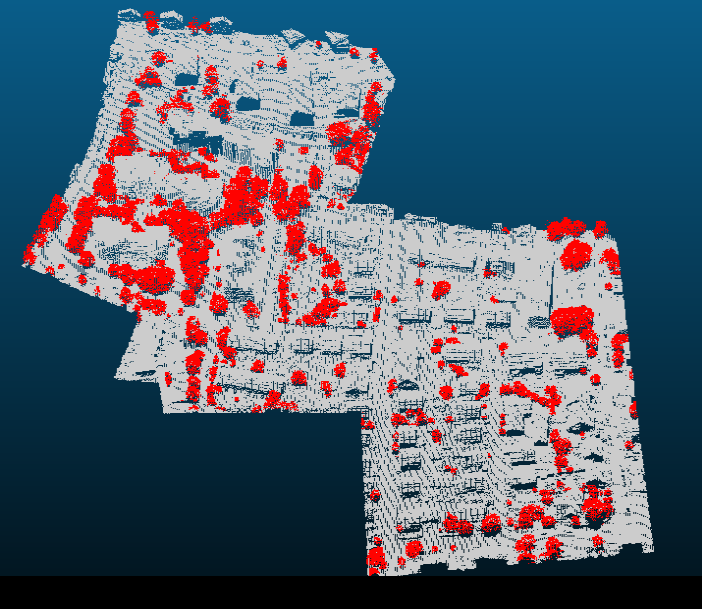}}\hfill
\subfigure[Segmentation Results- Heatmap corresponding to confidence in presence of tree: Red(High Confidence) to Blue(Low Confidence).]{\includegraphics[width=0.23\textwidth]{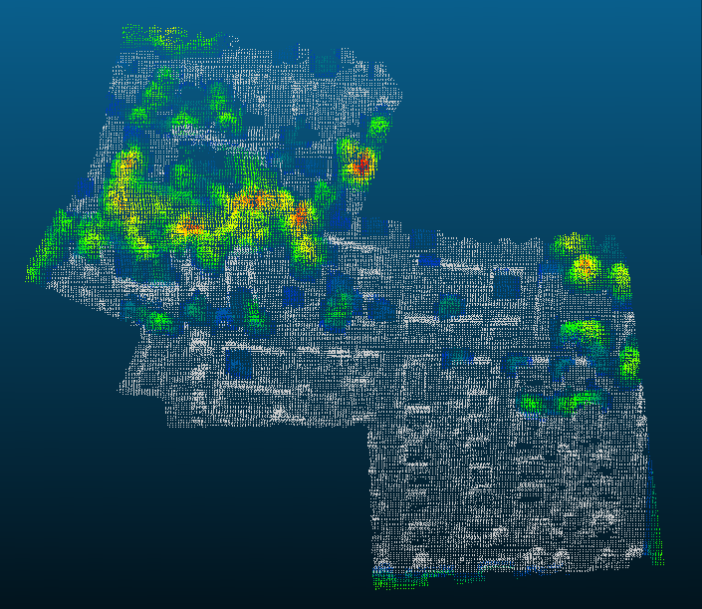}}
\caption[TreeNet Segmentation Results]{TreeNet Segmentation Results.}
\label{fig:tree_seg}
\end{figure}

% \begin{figure}[]
% \centering
% \subfigure[Voxel Resolution 4]{\includegraphics[width=.2\textwidth]{images/missed_tree_voxel}\label{fig:low_res}}\hfill
% \subfigure[Voxel Resolution 5]{\includegraphics[width=.2\textwidth]{images/missed_tree_voxel_5}\label{fig:high_res}}
% \caption[]{Effect of voxel resolution with tree voxels in green and non-tree voxels in red}
% \label{fig:vox_res}
% \end{figure}

% Different network topologies were tested and

Key results on the ISPRS test dataset are given in Table \ref{tab:aerial_seg}. The 'Val Accuracy' column gives the classification accuracy on the held-out validation set.

As can be seen from the results, the CNNs trained on the Dublin dataset performed worse than those trained on the Montreal dataset. This difference can be explained by the size of the datasets: the Dublin dataset covered only a \SI{2}{\kilo\metre\squared} radius in the city whereas the Montreal dataset was an area over \SI{775}{\kilo\metre\squared}. CNNs are prone to overfitting in small datasets and are able to learn more robust features with large datasets, hence the performance difference between the two datasets.

Another interesting result to note is that the TreeNet achieved better results when trained on a weakly labelled dataset (Montreal) than on the manually annotated dataset (ISPRS training). Intuitively, the latter should give better results since it was collected around the same time and the same location as the test set, which should have similar statistics. We believe this  can be explained by a couple of factors. Firstly, similar to the Dublin dataset, the ISPRS dataset was fairly small and the Montreal dataset, by virtue of its size, enabled better generalisation. Secondly, even though the downsampling caused a loss in detail, it also made the different datasets more uniform
% , reducing the disparity that might have been caused by different sensors and locations.

%  All networks had two convolutional layers with leaky ReLU non-linearity~\cite{Maas2013} and dropout, followed by two FC layers. 

The effect of corrupting the input data by dropping random voxels was evaluated and the results are given in Table \ref{tab:aerial_seg}. 20\% of the input dataset was corrupted by removing voxels randomly with probability of 0.5. It can be seen that this improved the accuracy by a small percentage. We believe this was due to the fact that randomising the input allows the network to learn more robust features and prevents overfitting.

Since the negative training samples were generated randomly from the training set, we also tested the effect of having a minimum number of occupied voxels in these samples, which is represented by the column 'Min Neg Samples' in Table \ref{tab:aerial_seg}.

The most accurate tree segmentation results on the ISPRS dataset are visualised in Figure \ref{fig:tree_seg} along with the ground truth labels. The results from the proposed methodology suggest that the technique is effective and promising, though not matching the state-of-the-art results. On further analysis of the results, it can be seen that the CNN was able to identify all the large trees but misses the smaller ones. This problem seems to occur due to the voxelisation process when the data is downsampled since  some of the smallest trees and bushes occupy a very small number of voxels, making it difficult for them to identify. The best results were achieved when the input data was corrupted by randomly dropping voxels during the training process, improving the generalisation of the network.

% As can be seen in Figure \ref{fig:low_res}, some of the smallest trees and bushes occupy a very small number of voxels, making it difficult for them to be identified. A number of different techniques have been tried in order to deal with this issue, including increasing voxel resolution, with limited success as shown in Figure \ref{fig:high_res}, and applying different jittering and data corruption techniques. The best results were achieved when the input data was corrupted by randomly dropping voxels during the training process, making the network robust to overfitting.

\subsection{sPointnet++ for Segmentation}

\begin{table}[]
\caption{sPointnet++ results with varying input types}
\label{tab:pointnet}
% \begin{adjustbox}{width=\textwidth}
\begin{tabular}{|l|l|l|l|l|}
\hline
\textbf{Input Type} & \textbf{Data Type} & \textbf{\(p\)} & \textbf{\(r\)} & \textbf{F\(_{score}\)} \\ \hline
x, y, z & Trees & 52 & 69.5 & 59.5 \\ \hline
x, y, z, i & Trees & 51.3 & 87.3 & 64.7 \\ \hline
x, y, z, i, ir, r, g & Trees & 79.5 & 84.9 & 82.1 \\ \hline
% x, y, z\(_{dtm}\), i, ir, r, g & Trees & 77 & 83.7 & 80.2 \\ \hline
x, y, z & Vegetation & 52.2 & 76.5 & 62 \\ \hline
x, y, z, i & Vegetation & 64.9 & 85.9 & 73.9 \\ \hline
x, y, z, i, ir, r, g & Vegetation & 82.9 & 85.4 & 84.1 \\ \hline
% x, y, z\(_{dtm}\), i, ir, r, g & Vegetation & 82.9 & 85.4 & 84.1 \\ \hline

\end{tabular}
% \end{adjustbox}
\end{table}

\begin{figure}[]
\centering
\subfigure[]{\includegraphics[width=.23\textwidth]{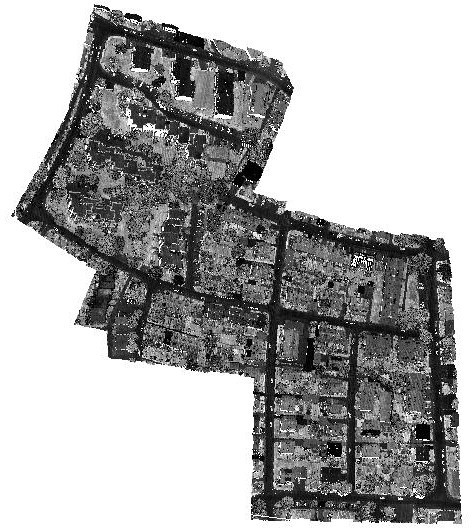}}\hfill
\subfigure[]{\includegraphics[width=.23\textwidth]{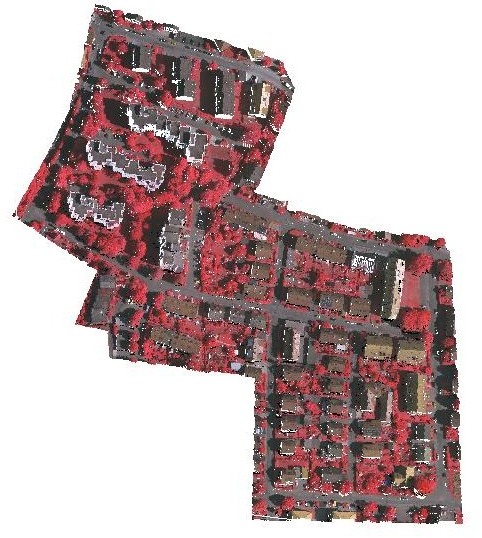}}
\caption[]{LiDAR data with point intensity in greyscale (left) and combined with spectral information (right).}
\label{fig:lidar_colour}
\end{figure}

\begin{figure}[h]
\centering
{\includegraphics[width=.4\textwidth]{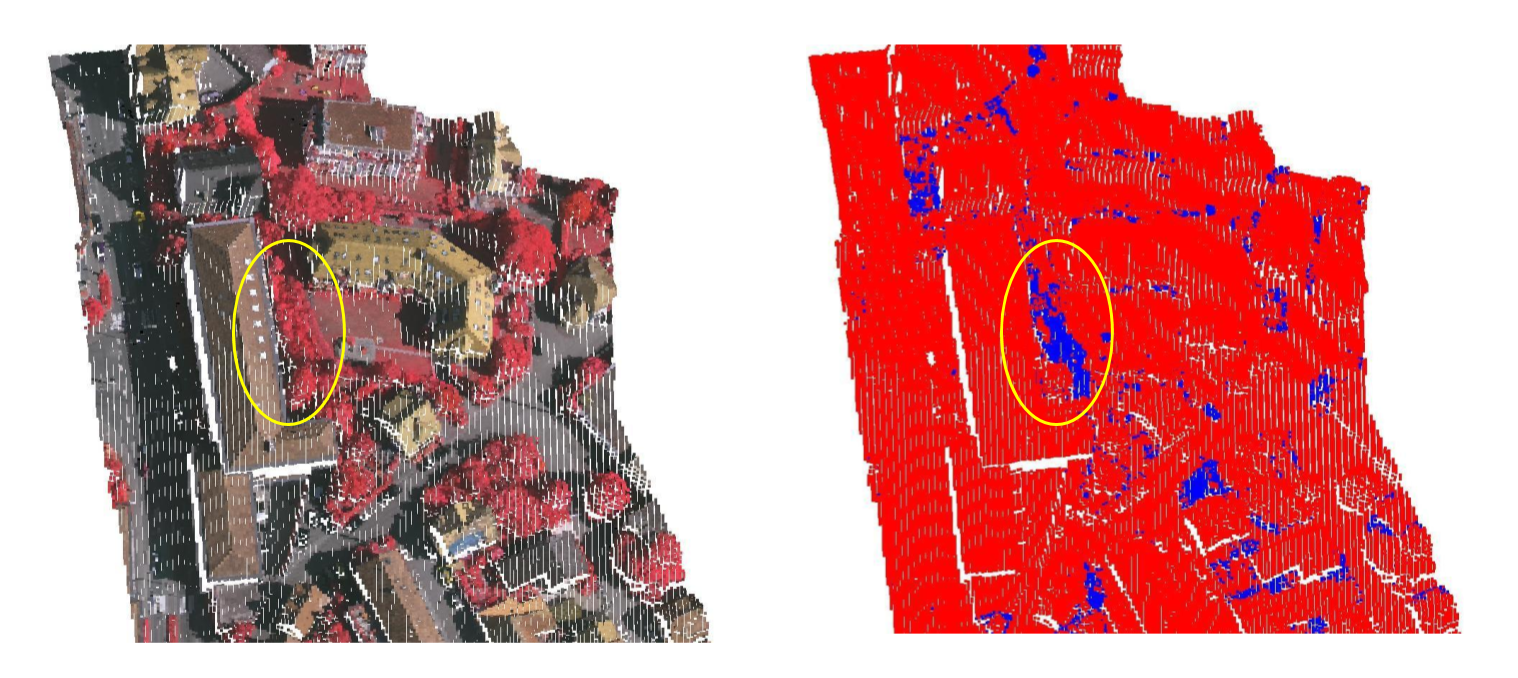}}\hfill
\caption[]{Area always incorrectly identified by network as tree outline in yellow in spectral image(left) and LiDAR point cloud(right). The LiDAR data is coloured red for all correctly labelled points and blue for all incorrectly labelled ones.}
\label{fig:error_image}
\end{figure}

The results of the TreeNet are limited by the voxelisation process, hence Pointnet++ was adapted to work directly on point clouds.

\begin{table*}[]
\caption{Comparison of Results on the ISPRS Benchmark Dataset}
\label{tab:isprs_comp}
\begin{adjustbox}{width=\textwidth}
\begin{tabular}{|l|l|l|l|l|l|}
\hline
\textbf{Method Name} & \textbf{Input Type} & \textbf{Method Description} & \textbf{Precision} & \textbf{Recall} & \textbf{F1} \\ \hline
IIS\_7~\citep{Ramiya2016} & LiDAR + Orthophoto (XYZRGB)  & Supervoxel Segmentation + Different ML Classifiers & 84 & 68.8 & 75.6 \\ \hline

WhuY3~\citep{Yang2017a} & Feature Images dervied from LiDAR point cloud & 2D CNN & 77.5 & 78.5 & 78.0 \\ \hline
% HM1 &  &  & 77.9 & 82.6 & 80.2 \\ \hline
% UM &  & One vs One Genetic Algorithm & 71.8 & 85.2 & 77.9 \\ \hline
RIT\_1~\citep{Yousefhussien2018} & LiDAR points, Spectral Image & Pointnet & 86.0 & 79.3 & 82.5 \\ \hline
NANJ2~\cite{Zhao2018} & LiDAR attributes, Geomtrical Features, DTM, Spectral Image & Multiscale 2D CNN with Bagged Decision Trees & 88.3 & 77.5 & 82.6 \\ \hline
% K\_LDA~\citep{Blomley2017} &  &  & 57.5 & 72.8 & 64.2 \\ \hline
LUH & LiDAR attributes, Geometrical profiles, Textural properties  & Two Layer CRF & 87.4 & 79.1 & 83.1 \\ \hline
% TreeNet (Ours) & Voxelised LiDAR points & 3D CNN & 60.1  & 64.4 & 62.2 \\ \hline
sPointnet++ (Ours) & LiDAR, Spectral Image & Point cloud based CNN & 79.5 & 84.9 & 82.1  \\ \hline
\end{tabular}
\end{adjustbox}
\end{table*}

\begin{figure*}[]
\centering
\subfigure[]{\frame{\includegraphics[width=.3\textwidth]{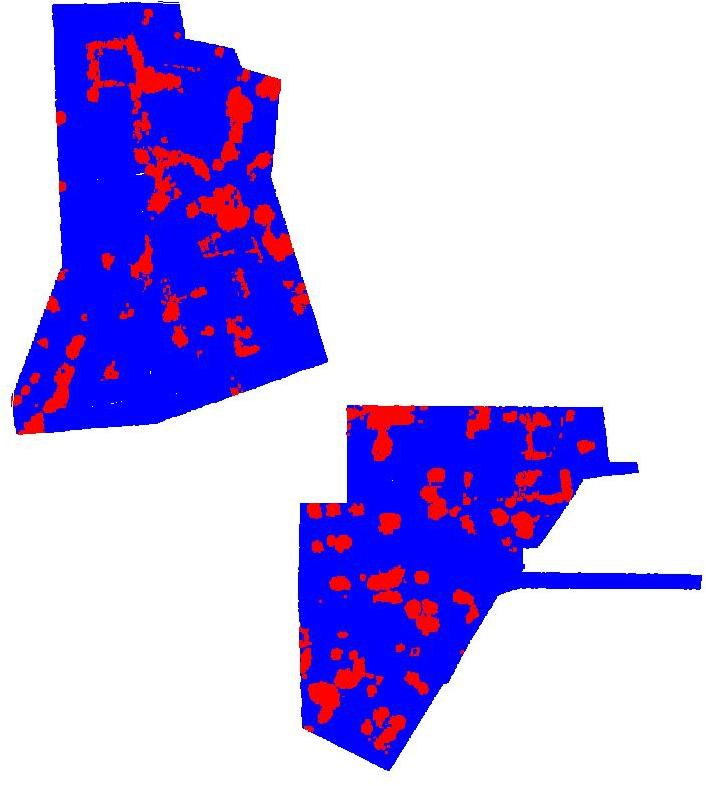}}}
\subfigure[]{\frame{\includegraphics[width=.3\textwidth]{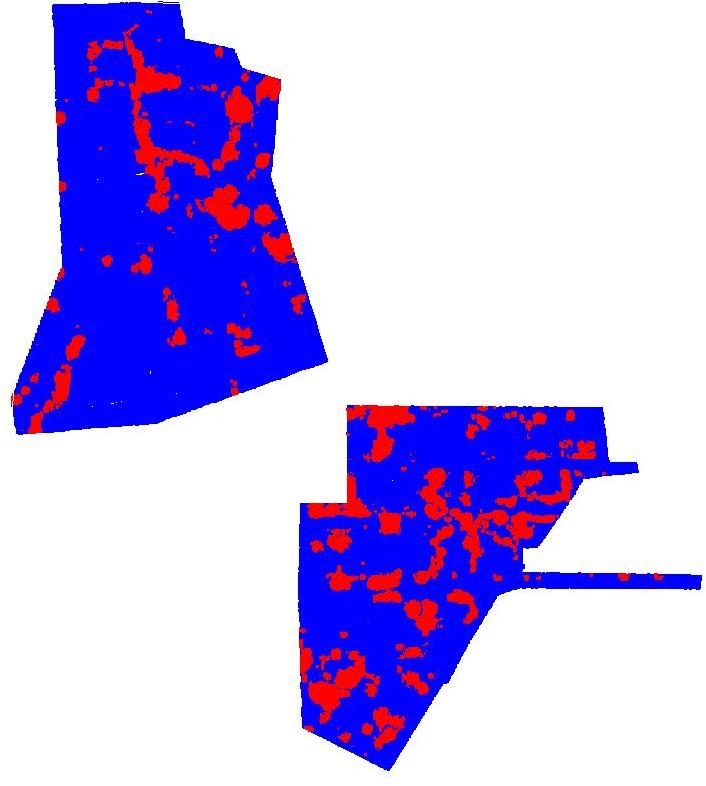}}}
\subfigure[]{\frame{\includegraphics[width=.3\textwidth]{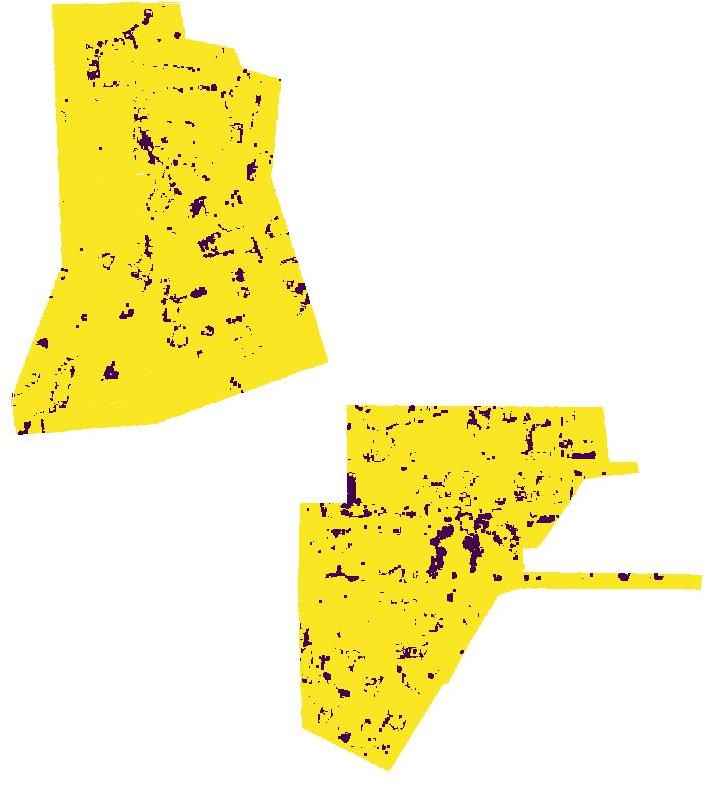}}}
\caption[]{Visualisation of sPointnet++ output; Left: Ground Truth with trees in red and non-trees in blue, Middle: Output of network (same colour scheme as previously), Right: Error voxels in Purple, correct voxels in Yellow. }
\label{fig:rgb_spectral_out}
\end{figure*}

% \todo{cite Yang2017a- they show that 1m is a good radius, Blomley2017 also has radius of 20m for height values, Zhao2018 uses multiscale upto 9.8 m}

The network was trained with different types of input to see the effect of different features. The first type of input was the x, y, z point coordinates, with the x and y inputs normalised per block. For the second input type, the intensity of each point was normalised between 0 and 1 and was concatenated to the point coordinates. For the third input type, the spectral information was added to the point cloud by interpolating the IR-R-G values from the 2D infrared image provided in the ISPRS dataset to the 3D point cloud. 

The results of the experiments are summarised in Table \ref{tab:pointnet}. It can be seen that with only point based inputs, the network was able to recognise 70\% of the trees correctly but its precision was low and it falsely identified a number of roofs and shrubs as tree points. The addition of the point intensity value increased the number of trees the network could identify but it still struggled with roofs and shrubs. 

To test how well the network was able to identify medium to high vegetation, we also marked all shrubs as positive samples; this gave better results. As can be seen from Table \ref{tab:pointnet}, the precision in this case was much higher as the shrubs identified as false positives in the previous experiments were now correct and the network had more positive samples to learn from.

Finally, the inclusion of the spectral information along with the point cloud data provided the best results with the top \(F_{score}\) of 82.1\% (mean 80.48, std dev. 0.877). This was due to the fact that spectral data provides highly discriminating information for identifying vegetation as can be seen in Figure \ref{fig:lidar_colour}, allowing the network to distinguish roofs from vegetation. 

Most misclassifications were shrubs and fences being identified as trees and some points at the edges of trees were missed by the network. The common errors made by all networks are shown in Figure \ref{fig:error_image}, where the area is identified as high vegetation but in the ground truth has been marked as low vegetation. The output from the network is visualised in Figure \ref{fig:rgb_spectral_out} along with the incorrectly identified points.

% This is possibly due to the fact that the ground elevation changes abruptly in that region. This error could be accounted for with the addition of a Digital Terrain Model for normalising the points along the z-axis.

We have compared the performance of sPointNet++ on the ISPRS benchmark with other methods in Table \ref{tab:isprs_comp}. The best results on the ISPRS benchmark are only marginally better than that of sPointNet++ but they are a lot more algorithmically intensive. Both LUH and NANJ2 are dependant on a complex preprocessing pipeline to generate geometric features and LUH further uses two independent conditional random fields (CRF) to aggregate the points.  RIT\_1 averages the results from multiple scales and hence requires multiple passes through their CNN to get multiscale features with \SI{3.7}{\second} for 412k points. In contrast, sPointNet++ requires minimal preprocessing of the dataset since it operates directly on the point cloud with spectral information. It also encodes hierarchical information with skip links and only requires a single pass through the CNN with \SI{2.8}{\second} for the same data, making it significantly more efficient than the state-of-the-art methods on the ISPRS benchmark.

\section{Conclusions}

In this paper, we have developed and shown both traditional methods and deep learning networks for identifying trees in LiDAR data with disparate resolutions. 

The proposed MultiReturn method works on high density LiDAR datasets and utilises the \textit{number of returns} LiDAR attribute to identify trees. It achieved almost 90\% accuracy on the Dublin dataset. Since this method does not scale well to low point-density datasets, we proposed a 3D CNN-based TreeNet to work with a low resolution 3D voxel grid. TreeNet was able to identify large trees in low density datasets but was unable to distinguish small trees due to loss of resolution in the voxelisation process. However, it did show good generalisation capability, since we were able to train on one dataset and test on a completely different dataset (different location, scanning hardware, point cloud statistics, etc).

We also proposed the scaled PointNet++, sPointNet++, which works on spectral data combined with aerial point cloud and does not require voxelisation.  The differences in using only point cloud data as compared to point clouds combined with spectral data have been analysed. We achieved comparable results to the state-of-the-art methods in tree identification using point clouds with an \(F_{score}\) of 82.1\% with a significantly more efficient pipeline. 

This work could be improved in several aspects. Ensembles of different models could be used to improve the performance. A larger annotated dataset would also improve performance for sPointNet++, further helping it generalise well. This work could also be extended to multi-class segmentation problems.

\section*{Acknowledgment}

The authors wish to thank the anonymous reviewers for their extremely valuable comments. Part of this work was done while A. Gupta was visiting the Intel Corporation funded by the HiPEAC4 Network of Excellence under the EU H2020 programme, grant agreement number 687698.

\ifCLASSOPTIONcaptionsoff
  \newpage
\fi



%\bibliographystyle{IEEEtran}


% \end{thebibliography}
\newpage

\bibliography{main}


% \end{IEEEbiography}


% % insert where needed to balance the two columns on the last page with
% %\newpage



%\vfill


\end{document}